\documentclass[10pt,twocolumn,letterpaper]{article}

\usepackage[pagenu mbers]{cvpr} %

\usepackage{graphicx}
\usepackage{framed,multirow}
\usepackage{amssymb}
\usepackage{latexsym}
\usepackage{algorithmic}
\usepackage{algorithm}
\usepackage{amsmath,amsfonts}
\usepackage{lineno}
\usepackage{amsmath}
\usepackage{amssymb}
\usepackage{booktabs}
\usepackage{cite}
\usepackage{amsmath,amssymb,amsfonts}
\usepackage{algorithmic}
\usepackage{graphicx}
\usepackage{amsmath}
\usepackage{bbding}
\usepackage{booktabs}
\usepackage{graphicx}
\usepackage{cite}
\usepackage{times}
\usepackage{dsfont}
\usepackage{amssymb}
\usepackage{multirow}
\usepackage{textcomp}

\usepackage[normalem]{ulem}
\usepackage{url}
\usepackage{dirtytalk}
\usepackage{algorithm,algorithmic}

\usepackage[pagebackref,breaklinks,colorlinks]{hyperref}
\usepackage[accsupp]{axessibility}

\usepackage[capitalize]{cleveref}
\crefname{section}{Sec.}{Secs.}
\Crefname{section}{Section}{Sections}
\Crefname{table}{Table}{Tables}
\crefname{table}{Tab.}{Tabs.}

\def\cvprPaperID{5588} %
\def\confName{CVPR}
\def\confYear{2022}

\begin{document}

\title{ACPL: Anti-curriculum Pseudo-labelling for \\ Semi-supervised Medical Image Classification}

\author{
\parbox{0.7\linewidth}{\centering $\quad$ Fengbei Liu\textsuperscript{\rm 1}\thanks{First two authors contributed equally to this work.}  $\quad$ Yu Tian\textsuperscript{\rm 1}\footnotemark[1] $\quad$    Yuanhong Chen\textsuperscript{\rm 1} $\quad$ Yuyuan Liu\textsuperscript{\rm 1}$\newline$  $\quad$  Vasileios Belagiannis\textsuperscript{\rm 2} $\quad$ $\quad$ $\quad$ Gustavo Carneiro\textsuperscript{\rm 1} $\newline$   
\textsuperscript{\rm 1} Australian Institute for Machine Learning, University of Adelaide \\
\textsuperscript{\rm 2} Universit\"at Ulm, Germany} 
}

\maketitle

\begin{abstract}
Effective semi-supervised learning (SSL) in medical image analysis (MIA) must address two challenges: 1) work effectively on both multi-class (e.g., lesion classification) and multi-label (e.g., multiple-disease diagnosis) problems, and 2) handle imbalanced learning (because of the high variance in disease prevalence).
One strategy to explore in SSL MIA is based on the pseudo labelling strategy, but it has a few shortcomings.
Pseudo-labelling has in general lower accuracy than consistency learning, it is not specifically design for both multi-class and multi-label problems, and it can be challenged by imbalanced learning. 
In this paper, unlike traditional methods that select confident pseudo label by threshold, we propose a new SSL algorithm, called
\textit{anti-curriculum pseudo-labelling (ACPL)}, which introduces novel techniques to select informative unlabelled samples, improving training balance and allowing the model to work for both multi-label and multi-class problems, and to estimate pseudo labels by an accurate ensemble of classifiers (improving pseudo label accuracy).
We run extensive experiments to evaluate ACPL on two public medical image classification benchmarks: Chest X-Ray14 for thorax disease multi-label classification and ISIC2018 for skin lesion multi-class classification. Our method outperforms previous SOTA SSL methods on both datasets\footnote{Supported by Australian Research Council through grants DP180103232 and FT190100525.}\footnote{Code is available at https://github.com/FBLADL/ACPL}. 
\end{abstract}

\section{Introduction}
\label{sec:introduction}

\begin{figure}
    \centering
    \begin{subfigure}[b]{0.48\textwidth}
    \includegraphics[width=\linewidth]{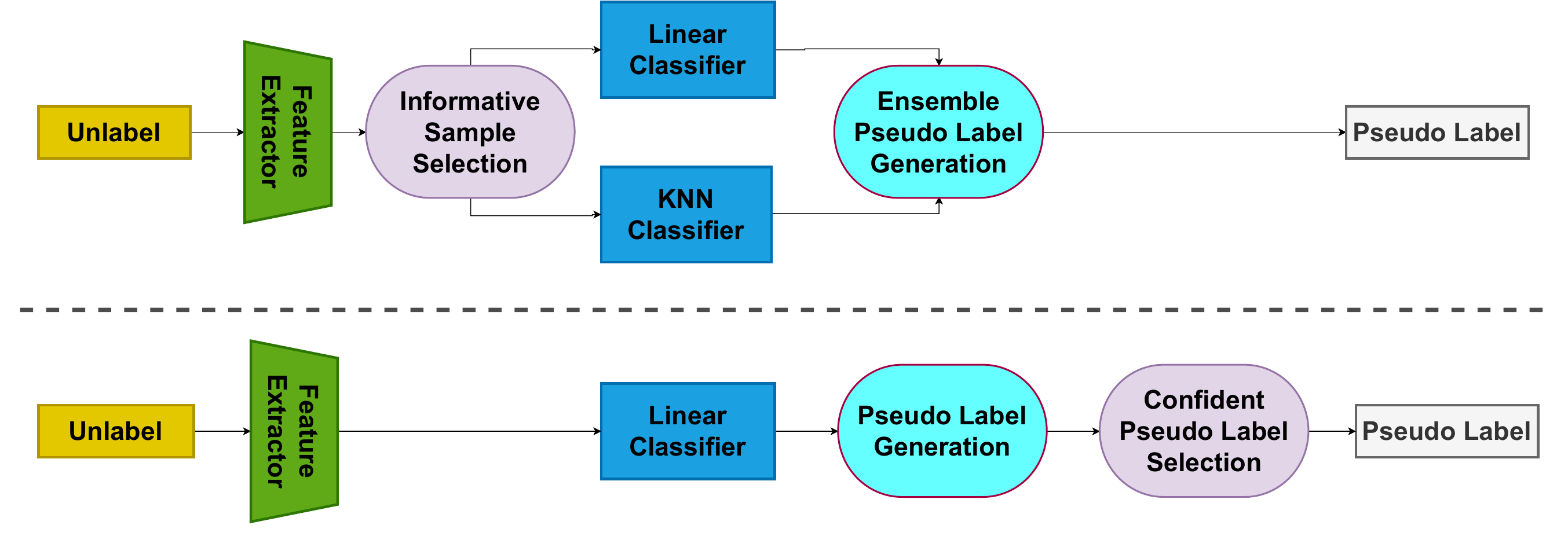}
    \caption{Diagram of our ACPL (top) and traditional pseudo-label SSL (bottom)}
    \label{fig:general_diagram}
    \end{subfigure}
    \begin{subfigure}[b]{0.48\textwidth}
    \includegraphics[width=\linewidth]{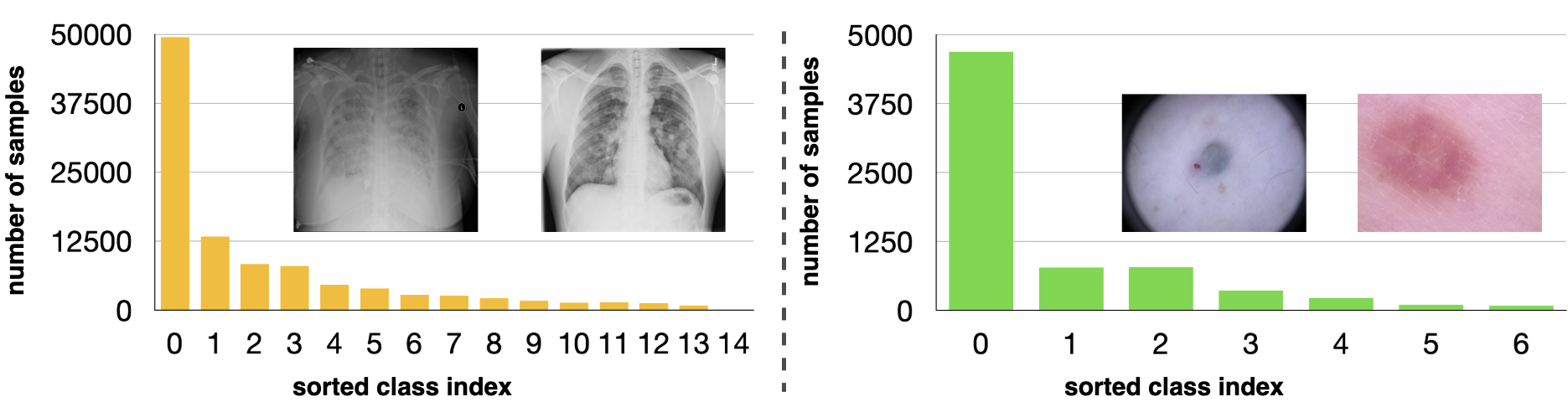}
    \caption{Imbalanced distribution on multi-label Chest X-ray14~\cite{wang2017chestx} (left) and multi-class ISIC2018~\cite{tschandl2018ham10000} (right)}
    \label{fig:distribution}
    \end{subfigure}
    \caption{ In (a), we show diagrams of the proposed ACPL (top) and the traditional pseudo-label SSL (bottom) methods, and (b) displays histograms of images per label for the multi-label Chest X-ray14~\cite{wang2017chestx} (left) and multi-class ISIC2018~\cite{tschandl2018ham10000} (right). 
    }
    \label{fig:first_figure}
\end{figure}

Deep learning has shown outstanding results in medical image analysis (MIA)~\cite{litjens2017survey,tian2020few,tian2021constrained}.
Compared to computer vision, the labelling of MIA training sets by medical experts is significantly more expensive, resulting in low availability of labelled images, but the high availability of unlabelled images from clinics and hospitals databases can be explored in the modelling of deep learning classifiers.
Furthermore, differently from computer vision problems that tend to be mostly multi-class and balanced, MIA has a number of multi-class (e.g., a lesion image of a single class) and multi-label (e.g., an image from a patient can contain multiple diseases) problems, where both problems usually contain severe class imbalances because of the variable prevalence of diseases (see Fig.~\ref{fig:first_figure}-(b)).
Hence, MIA semi-supervised learning (SSL) methods need to be flexible enough to work with multi-label and multi-class problems, in addition to handle imbalanced learning.

State-of-the-art (SOTA) SSL approaches are usually based on the consistency learning of unlabelled data~\cite{berthelot2019mixmatch,berthelot2019remixmatch,sohn2020fixmatch} and self-supervised pre-training~\cite{liu2021self}.
Even though consistency-based methods show SOTA results on multi-class SSL problems, pseudo-labelling methods have shown better results for multi-label SSL problems~\cite{rizve2020defense}.
Pseudo-labelling methods provide labels to confidently classified unlabelled samples that are used to re-train the  model~\cite{lee2013pseudo}.
One issue with pseudo-labelling SSL methods is that the confidently classified unlabelled samples represent the least informative ones~\cite{settles2009active} that, for imbalanced problems, are likely to belong to the majority classes.
Hence, this will bias the classification toward the majority classes and most likely deteriorate the classification accuracy of the minority classes.
Also, selecting confident pseudo-labelled samples is challenging in multi-class, but even more so in multi-label problems. 
Previous papers~\cite{rizve2020defense,aviles2019graphx} use a fixed threshold for all classes, but a class-wise threshold
that addresses imbalanced learning and correlations between classes in multi-label problems would enable more accurate pseudo-label predictions.
However, such class-wise threshold is hard to estimate without knowing the class distributions or if we are dealing with a multi-class or multi-label problem.
Furthermore, using the model output for the pseudo-labelling process can also cause confirmation bias~\cite{arazo2020pseudo}, whereby 
the assignment of incorrect pseudo-labels  will increase the model confidence in those incorrect predictions, and consequently decrease the model accuracy.

In this paper, we propose the \textbf{anti-curriculum pseudo-labelling (ACPL)}, which addresses multi-class and multi-label imbalanced learning SSL MIA problems.
First, we introduce a new approach to select the most informative unlabelled images to be pseudo-labelled.
This is motivated by our argument that there exists a distribution shift between unlabelled and labelled samples for SSL. An effective learning curriculum must focus on informative unlabelled samples that are located as far as possible from the distribution of labelled samples. As a result, these informative samples are likely to belong to the minority classes in MIA imbalanced learning problems. Selecting these informative samples  will naturally balance the training process and, given that they are selected before the pseudo-labelling process, we eliminate the need for estimating a class-wise classification threshold, facilitating our model to work well on multi-class and multi-label problems.
The information content measure of an unlabelled sample is computed with our proposed  cross-distribution sample informativeness that outputs how close an unlabelled sample is from the set of labelled anchor samples (anchor samples are highly informative labelled samples). 
Second, we introduce a new pseudo-labelling mechanism, called informative mixup, which combines the model classification with a K-nearest neighbor (KNN) classification guided by sample informativeness to improve prediction accuracy and mitigate confirmation bias. 
Third, we propose the anchor set purification method
 that selects the most informative pseudo-labelled samples to be included in the labelled anchor set to improve the pseudo-labelling accuracy of the KNN classifier in later training stages.

To summarise, our ACPL approach selects highly informative samples for pseudo-labelling (addressing MIA imbalanced classification problems and allowing multi-label multi-class modelling) and uses an ensemble of classifiers to produce accurate pseudo labels (tackling confirmation bias to improve classification accuracy), where the main technical contributions are:
\begin{itemize}
    \item A novel information content measure to select informative unlabelled samples named \textbf{cross-distribution sample informativeness};
    \item A new pseudo-labelling mechanism, called \textbf{informative mixup}, which generates pseudo labels from an ensemble of deep learning and KNN classifiers; and
    \item A novel method, called
    \textbf{anchor set purification (ASP)}, to select informative pseudo-labelled samples to be included in the labelled anchor set to improve the pseudo-labelling accuracy of the KNN classifier.
\end{itemize}
We evaluate ACPL on two publicly available medical image classification datasets, namely the Chest X-Ray14 for thorax disease multi-label classification~\cite{wang2017chestx} and the ISIC2018 for skin lesion multi-class classification~\cite{tschandl2018ham10000, codella2019skin}. Our method outperforms the current SOTA methods in both datasets.

\section{Related Work}
\label{sec:related_work}

We first review consistency-based and pseudo-labelling SSL methods.
Then, we discuss the curriculum and anti-curriculum learning literature for fully and semi-supervised learning and present relevant SSL MIA methods.

\textbf{Consistency-based SSL} optimises the classification prediction of labelled images and minimises the prediction outputs of different views of unlabelled images, where these views are obtained from different types of image perturbations, such as spatial/temporal~\cite{laine2016temporal,tarvainen2017mean}, adversarial~\cite{miyato2018virtual}, or data augmentation~\cite{berthelot2019mixmatch,sohn2020fixmatch, berthelot2019remixmatch}.
The performance of the consistency-based methods can be further improved with self-supervised pre-training~\cite{liu2021self}.
Even though consistency-based SSL methods show SOTA results in many benchmarks~\cite{sohn2020fixmatch}, they depend on a careful design of perturbation functions that requires domain knowledge and would need to be adapted to each new type of medical imaging. Furthermore, Rizve et al.~\cite{rizve2020defense} show that pseudo-labelling SSL methods are more accurate for multi-label problems.

\textbf{Pseudo-labelling SSL} methods~\cite{rizve2020defense,shi2018transductive,xie2020self,cascante2020curriculum} train a model with the available labelled data, estimate the pseudo labels of unlabelled samples classified with high confidence~\cite{lee2013pseudo}, then take these pseudo-labelled samples to re-train the model.
As mentioned above in Sec.~\ref{sec:introduction} pseudo-label SSL approaches can bias classification toward the majority classes in imbalanced problems, is not seamlessly adaptable to multi-class and multi-label problems, and can also lead to confirmation bias.
We argue that the improvement of pseudo-labelling SSL methods depends on the selection of informative unlabelled samples to address the majority class bias and the adaptation to multi-class and multi-label problems, and an accurate pseudo-labelling mechanism to handle confirmation bias, which are two points that we target with this paper.

The selection of training samples based on their information content has been studied by fully supervised \textbf{curriculum and anti-curriculum learning methods}~\cite{wu2021when}.
Curriculum learning focuses on the easy samples in the early training stages and gradually includes the hard samples in the later training stages, where easy samples~\cite{bengio2009curriculum,kumar2010self,jiang2015self} are usually defined as samples that have small losses during training, and hard samples tend to have large losses. 
On the other hand, anti-curriculum focuses of the hard samples first and transitions to the easy samples later in the training~\cite{jiang2019accelerating,kawaguchi2020ordered}. 
The methods above have been designed to work in fully supervised learning. 
Cascante et al.~\cite{cascante2020curriculum} explored a pseudo labelling SSL method based on curriculum learning, but we are not aware of SSL methods that explore anti-curriculum learning. %
Since we target accurate SSL of imbalanced multi-class and multi-label methods, we follow anti-curriculum learning that pseudo-labels the most informative samples which are likely to belong to the minority classes (consequently, helping to balance the training) and enable the selection of samples without requiring the estimation of a class-wise classification threshold (enabling a seamless adaptation to multi-class and multi-label problems).

The main benchmarks for SSL in MIA study the multi-label classification of chest X-ray (CXR) images~\cite{wang2017chestx,irvin2019chexpert} and multi-class classification of skin lesions~\cite{tschandl2018ham10000,codella2019skin}.
For \textbf{CXR SSL} classification, pseudo-labelling methods have  been explored~\cite{aviles2019graphx}, but SOTA results are achieved with consistency learning approaches~\cite{li2018semi,cui2019semi,liu2020semi,unnikrishnan2020semi,liu2021self}.
For \textbf{skin lesion SSL} classification, the current SOTA is also based on consistency learning~\cite{liu2020semi}, with pseudo-labelling  approaches~\cite{bai2017semi} not being competitive.
We show that our proposed pseudo-labelling method ACPL can surpass the consistency-based SOTA on both benchmarks, demonstrating the value of selecting highly informative samples for pseudo labelling and of the accurate pseudo labels from the ensemble of classifiers.
We also show that our ACPL improves the current computer vision SOTA~\cite{rizve2020defense} applied to MIA, demonstrating the limitation of computer vision methods in MIA and also the potential of our approach to be applied in more general SSL problems.

\section{Methods}
\label{sec:methods}
\begin{algorithm}
  \caption{Anti-curriculum Pseudo-labelling Algorithm\label{alg:acpl}}
  \begin{algorithmic}[1]
    \STATE \textbf{require: } Labelled set $\mathcal{D}_L$, unlabelled set $\mathcal{D}_U$, and number of training stages $T$
    \STATE \textbf{initialise} $\mathcal{D}_A = \mathcal{D}_L \text{, and } t=0$ 
    \STATE \textbf{warm-up train} {$p_{\theta_{t}}(\mathbf{x})$} \textbf{with}\\ 
    {$\theta_{t} = \arg\min_{\theta} 
    \frac{1}{|\mathcal{D}_L|}\sum_{(\mathbf{x}_i,\mathbf{y}_i) \in \mathcal{D}_L}
    \ell(\mathbf{y}_i,p_{\theta}(\mathbf{x}_i))$}
    \WHILE{$t < T$ \OR $|\mathcal{D}_U| \neq 0$}
    \STATE \textbf{build pseudo-labelled dataset using CDSI from~\eqref{eq:information_estimator} and IM from~\eqref{eq:pseudo_label}:}\\ 
    \vspace{-0.7cm}
    \begin{align*}
      \hspace{-1.2cm} \mathcal{D}_S = 
        \{(\mathbf{x},\tilde{\mathbf{y}})| &\mathbf{x} \in \mathcal{D}_U,  h(f_{\theta_t}(\mathbf{x}),\mathcal{D}_A) = 1 , \\
        & \tilde{\mathbf{y}}  =  g(f_{\theta_t}(\mathbf{x}),\mathcal{D}_A)
        \}
    \end{align*}
    \vspace{-0.7cm}
    
    \STATE \textbf{update anchor set with ASP from~\eqref{eq:anchor_set_insert}:}\\
    \vspace{-0.7cm}
    \begin{align*}
        \hspace{-0.6cm}\mathcal{D}_A = 
        &\mathcal{D}_A \bigcup  (\mathbf{x},\tilde{\mathbf{y}}) \text{, where} \\
        & (\mathbf{x},\tilde{\mathbf{y}}) \in \mathcal{D}_S \text{, and } a(f_{\theta_t}(\mathbf{x}),\mathcal{D}_U,\mathcal{D}_A)=1 
    \end{align*}
    \vspace{-0.7cm}
    \STATE {$t \leftarrow t + 1$}
    \STATE \textbf{optimise~\eqref{eq:optimisation} using } {$\mathcal{D}_L,\mathcal{D}_S$ \textbf{to obtain}  $p_{\theta_t}(\mathbf{x})$}
    \STATE \textbf{update labelled and unlabelled sets:} \\ {$\mathcal{D}_L \leftarrow \mathcal{D}_L \bigcup \mathcal{D}_S, \mathcal{D}_U \leftarrow \mathcal{D}_U \setminus \mathcal{D}_S$}
    \ENDWHILE
    \RETURN {$p_{\theta_t}(\mathbf{x})$}
  \end{algorithmic}
  \label{alg:acpl}
\end{algorithm}

To introduce our SSL method ACPL, assume that we have a small labelled training set $\mathcal{D}_L = \{ (\mathbf{x}_i,\mathbf{y}_i)\}_{i=1}^{|\mathcal{D}_L|}$, where $\mathbf{x}_i \in \mathcal{X} \subset \mathbb{R}^{H \times W \times C}$ is the input image of size $H \times W$ with $C$ colour channels, and $\mathbf{y}_i  \in \{0,1\}^{|\mathcal{Y}|}$ is the label with the set of classes denoted by $\mathcal{Y}=\{1,...,|\mathcal{Y}|\}$ (note that $\mathbf{y}_i$ is a one-hot vector for multi-class problems and a binary vector in multi-label problems). 
A large unlabelled training set $\mathcal{D}_U = \{ \mathbf{x}_i\}_{i=1}^{|\mathcal{D}_U|}$ is also provided, with $|\mathcal{D}_L| << |\mathcal{D}_U|$.
We assume the samples from both datasets are drawn from the same (latent) distribution.
Our algorithm also relies on the pseudo-labelled set $\mathcal{D}_S$ that is composed of pseudo-labelled samples classified as informative unlabelled samples, and an anchor set $\mathcal{D}_A$ that contains informative pseudo-labelled samples. 
The goal of ACPL is to learn a model $p_\theta:\mathcal{X} \to [0,1]^{|\mathcal{Y}|}$ parameterised by $\theta$ using the labelled, unlabelled, pseudo-labelled, and anchor datasets.

Below, in Sec.~\ref{sec:acpl_optimisation}, we introduce our ACPL optimisation that produces accurate pseudo labels to unlabelled samples following an anti-curriculum strategy, where highly informative unlabelled samples are selected to be pseudo-labelled at each training stage. 
In Sec.~\ref{sec:information_content}, we present the information criterion of an unlabelled sample, referred to as \textit{cross distribution sample informativeness (CDSI)}, based on the dissimilarity between the unlabelled sample and samples in the anchor set $\mathcal{D}_A$.
The pseudo labels for the informative unlabelled samples are generated using the proposed \textit{informative mixup (IM)} method (Sec.~\ref{sec:pseudo_label}) that mixes up the results from the model $p_{\theta}(.)$ and a $K$ nearest neighbor (KNN) classifier using the anchor set. 
At the end of each training stage, the anchor set is updated with the
\textit{anchor set purification (ASP)} method (Sec.~\ref{sec:asp}) that only keeps the
most informative subset of pseudo-labelled samples, according to the \textit{CDSI} criterion.

\begin{figure}
    \centering
    \includegraphics[width=\linewidth]{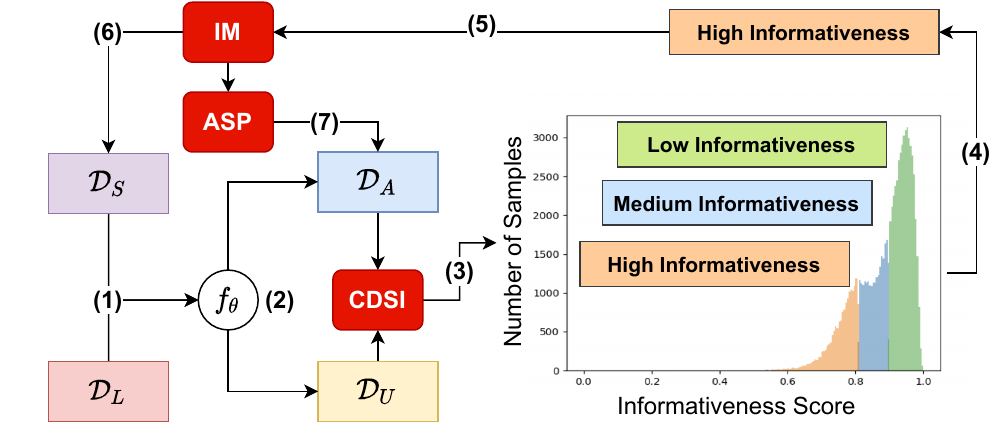}
    \caption{Anti-curriculum pseudo-labelling (ACPL) algorithm. 
    The algorithm is divided into the following iterative steps: 1) train the model with $\mathcal{D}_S$ and $\mathcal{D}_L$; 2) extract the features from the anchor and unlabelled samples; 3) estimate information content of unlabelled samples with CDSI from~\eqref{eq:informativeness} with anchor set $\mathcal{D}_A$; 4) partition the unlabelled samples into high, medium and low information content using~\eqref{eq:information_estimator}; 5) assign a pseudo label to high information content unlabelled samples with IM from~\eqref{eq:pseudo_label}; 6) update $\mathcal{D}_S$ with new pseudo-labelled samples; and 7) update $\mathcal{D}_A$ with ASP in~\eqref{eq:anchor_set_insert}.} 
    \label{fig:ACPL}
\end{figure}

\subsection{ACPL Optimisation}
\label{sec:acpl_optimisation}

Our ACPL optimisation, described in Alg.~\ref{alg:acpl} and depicted by Fig.~\ref{fig:ACPL}, starts with 
a warm-up supervised training of the parameters of the model $p_{\theta}(.)$ 
using only the labelled set $\mathcal{D}_L$.
For the rest of the training, we use the sets of labelled and unlabelled samples, $\mathcal{D}_L$ and $\mathcal{D}_U$, and update the pseudo-labelled set $\mathcal{D}_S$ and the  anchor set $\mathcal{D}_A$ containing the informative unlabelled and pseudo-labelled samples, where $\mathcal{D}_S$ start as an empty set 
and $\mathcal{D}_A$ starts with the samples in $\mathcal{D}_L$. 
The optimisation iteratively minimises the following cost function: %
\begin{equation}
\begin{split}
\ell_{ACPL}(\theta, \mathcal{D}_L,\mathcal{D}_S) & =\frac{1}{|\mathcal{D}_L|}\sum_{(\mathbf{x}_i,\mathbf{y}_i)  \in \mathcal{D}_L}
\ell(\mathbf{y}_i,p_{\theta}(\mathbf{x}_i))\\
& + \frac{1}{|\mathcal{D}_S|}\sum_{(\mathbf{x}_i,\tilde{\mathbf{y}}_i) \in \mathcal{D}_S}
\ell(\tilde{\mathbf{y}}_i,p_{\theta}(\mathbf{x}_i)),%
\end{split}
\label{eq:optimisation}
\end{equation}
where $\ell(.)$ denotes a classification loss (\textit{e.g}., cross-entropy), $\theta$ is the model parameter, $\mathbf{y}_i$ is the ground truth, and $\tilde{\mathbf{y}}_i$ is the estimated pseudo label.
After optimising~\eqref{eq:optimisation}, the labelled and unlabelled sets are updated as $\mathcal{D}_L = \mathcal{D}_L \bigcup \mathcal{D}_S$ and 
$\mathcal{D}_U = \mathcal{D}_U \setminus \mathcal{D}_S$, and a new iteration of optimisation takes place.

\subsection{Cross Distribution Sample Informativeness (CDSI)}
\label{sec:information_content}

The function that estimates if an unlabelled sample has high information content is defined by
\begin{equation}
    h(f_{\theta}(\mathbf{x}),\mathcal{D}_A) =
\left\{\begin{array}{ll} 
1, & p_{\gamma}(\zeta=\text{high} | \mathbf{x}, \mathcal{D}_A) > \tau,\\
0, & \text{otherwise},
\end{array}
\right.	
    \label{eq:information_estimator}
\end{equation}
where $\zeta \in \mathcal{Z} = \{ \text{low}, \text{medium}, \text{high}\}$ represents the information content random variable, 
$\gamma = \{\mu_{\zeta},\Sigma_{\zeta},\pi_{\zeta} \}_{\zeta \in \mathcal{Z}}$ denotes the parameters of the model $p_{\gamma}(.)$,
and $\tau = \max\left \{ p_{\gamma}(\zeta=\text{low} | \mathbf{x}, \mathcal{D}_A),p_{\gamma}(\zeta=\text{medium} | \mathbf{x}, \mathcal{D}_A) \right \}$. The function
$p_{\gamma}(\zeta| \mathbf{x}, \mathcal{D}_A)$ can be decomposed into $p_{\gamma}(\mathbf{x} | \zeta, \mathcal{D}_A)p_{\gamma}(\zeta| \mathcal{D}_A)/p_{\gamma}(\mathbf{x} | \mathcal{D}_A)$, where
\begin{equation}
    p_{\gamma}(\mathbf{x} | \zeta, \mathcal{D}_A) =  n(d(f_{\theta}(\mathbf{x}),\mathcal{D}_A)|\mu_{\zeta},\Sigma_{\zeta}),
    \label{eq:gaussian}
\end{equation}
with $n(.;\mu_{\zeta},\Sigma_{\zeta})$ denoting a Gaussian function with mean $\mu_{\zeta}$ and covariance $\Sigma_{\zeta}$,
$p_{\gamma}(\zeta| \mathcal{D}_A)=\pi_{\zeta}$
representing the ownership probability of $\zeta$ (i.e., the weight of mixture $\zeta$), and $p_{\gamma}(\mathbf{x} | \mathcal{D}_A)$ being a normalisation factor.
The probability in~\eqref{eq:gaussian} is computed with the density of the unlabelled sample $\mathbf{x}$ with respect to the anchor set $\mathcal{D}_A$, as follows:
\begin{equation}
    d(f_{\theta}(\mathbf{x}),\mathcal{D}_A) = \frac{1}{K}\sum\limits_{\substack{(f_{\theta}(\mathbf{x}_A),\mathbf{y}_A) \in \\  \mathcal{N}(f_{\theta}(\mathbf{x}),\mathcal{D}_A)}} \frac{f_{\theta}({\mathbf{x}})^{\top}f_{\theta}({\mathbf{x}_A})} {\| f_{\theta}({\mathbf{x}}) \|_2 \| f_{\theta}({\mathbf{x}_A}) \|_2},
    \label{eq:informativeness}
\end{equation}
where $\mathcal{N}(f_{\theta}(\mathbf{x}),\mathcal{D}_A)$ represents the set of K-nearest neighbors (KNN) from
the anchor set $\mathcal{D}_A$ to the input image feature $f_{\theta}(\mathbf{x})$, with each element in the set $\mathcal{D}_A$ denoted by $(f_{\theta}(\mathbf{x}_A),\mathbf{y}_A)$.
The $F-$dimensional input image feature is extracted with
$f_{\theta}:\mathcal{X} \to \mathbb{R}^F$ from the model $p_{\theta}(.)$ with $p_{\theta}(\mathbf{x}) = \sigma(f_{\theta}(\mathbf{x}))$, where $\sigma(.)$ is the final activation function to produce an output in $[0,1]^{|\mathcal{Y}|}$.
The parameters $\gamma$ in~\eqref{eq:information_estimator} are estimated with the expectation-maximisation (EM) algorithm~\cite{dempster1977maximum}, every time after the anchor set is updated.

\subsection{Informative Mixup (IM)}

\label{sec:pseudo_label}

After selecting informative unlabelled samples with~\eqref{eq:information_estimator}, we aim to produce reliable pseudo labels for them. 
We can provide two pseudo labels for each unlabelled sample $\mathbf{x} \in \mathcal{D}_U$: the model prediction from $p_{\theta}(\mathbf{x})$, and the K-nearest neighbor (KNN) prediction using the anchor set, as follows:
\begin{equation}
    \begin{split}
        \tilde{\mathbf{y}}_{\text{model}}(\mathbf{x}) &= p_{\theta}(\mathbf{x}), \\
        \tilde{\mathbf{y}}_{\text{KNN}}(\mathbf{x}) &= \frac{1}{K}\sum_{(f_{\theta}(\mathbf{x}_A),\mathbf{y}_A) \in  \mathcal{N}(f_{\theta}(\mathbf{x}),\mathcal{D}_A)}  \mathbf{y}_A.
    \end{split}
    \label{eq:pseudo_label_types}
\end{equation}
$\mathbf{y}_A$ is the label of anchor set samples.
However, using any of the pseudo labels from~\eqref{eq:pseudo_label_types} can be problematic for model training. 
The pseudo label in $\tilde{\mathbf{y}}_{\text{model}}(\mathbf{x})$ can cause confirmation bias, and the reliability of $\tilde{\mathbf{y}}_{\text{KNN}}(\mathbf{x})$ depends on the size and representativeness of the initial labelled set to produce accurate classification. 
Inspired by MixUp~\cite{zhang2017MixUp}, we propose the \textbf{informative mixup} method that constructs the pseudo-labelling function $g(.)$ in~\eqref{eq:optimisation} with a linear combination of  $\tilde{\mathbf{y}}_{\text{model}}(\mathbf{x})$ and $\tilde{\mathbf{y}}_{\text{KNN}}(\mathbf{x})$ weighted by the density score from~\eqref{eq:informativeness}, as follows:
\begin{equation}
\begin{split}
    \tilde{\mathbf{y}}  = g(f_{\theta}(\mathbf{x}),\mathcal{D}_A) 
                      & = d(f_{\theta}(\mathbf{x}),\mathcal{D}_A) \times \tilde{\mathbf{y}}_{\text{model}}(\mathbf{x}) \\
                      & + (1-d(f_{\theta}(\mathbf{x}),\mathcal{D}_A)) \times \tilde{\mathbf{y}}_{\text{KNN}}(\mathbf{x}).
    \label{eq:pseudo_label}
\end{split}
\end{equation}
The informative mixup in~\eqref{eq:pseudo_label} is different from MixUp~\cite{zhang2017MixUp} because it combines the classification results of the same image from two models instead of the classification from the same model of two  images.
Furthermore, our informative mixup weights the the classifiers with the density score to reflect the trade-off between $\tilde{\mathbf{y}}_{\text{model}}(\mathbf{x})$ and $\tilde{\mathbf{y}}_{\text{KNN}}(\mathbf{x})$. 
Since informative samples are selected from a region of the anchor set with low feature density, the KNN prediction $\tilde{\mathbf{y}}_{\text{KNN}}(\mathbf{x})$ is less reliable than $\tilde{\mathbf{y}}_{\text{model}}(\mathbf{x})$, so by default, we should trust more  the model classification. 
The weighting between the two predictions in~\eqref{eq:pseudo_label} 
reflects this observation, where $\tilde{\mathbf{y}}_{\text{model}}(\mathbf{x})$ will tend have a larger weight given that $d(f_{\theta}(\mathbf{x}),\mathcal{D}_A)$ is usually larger than $0.5$, as displayed in Fig.~\ref{fig:ACPL} (see the informativeness score histogram at the bottom-right corner).
When the sample is located in a high-density region, we place most of the weight on the model prediction given that in such case, the model is highly reliable. On the other hand, when the sample is in a low-density region, we try to balance a bit more the contribution of both the model and KNN predictions, given the low reliability of the model.

\subsection{Anchor Set Purification (ASP)}
\begin{figure}
    \centering
    \includegraphics[clip,trim=0.0cm 0.0cm 5.0cm 0.0cm, width=\linewidth,page=3]{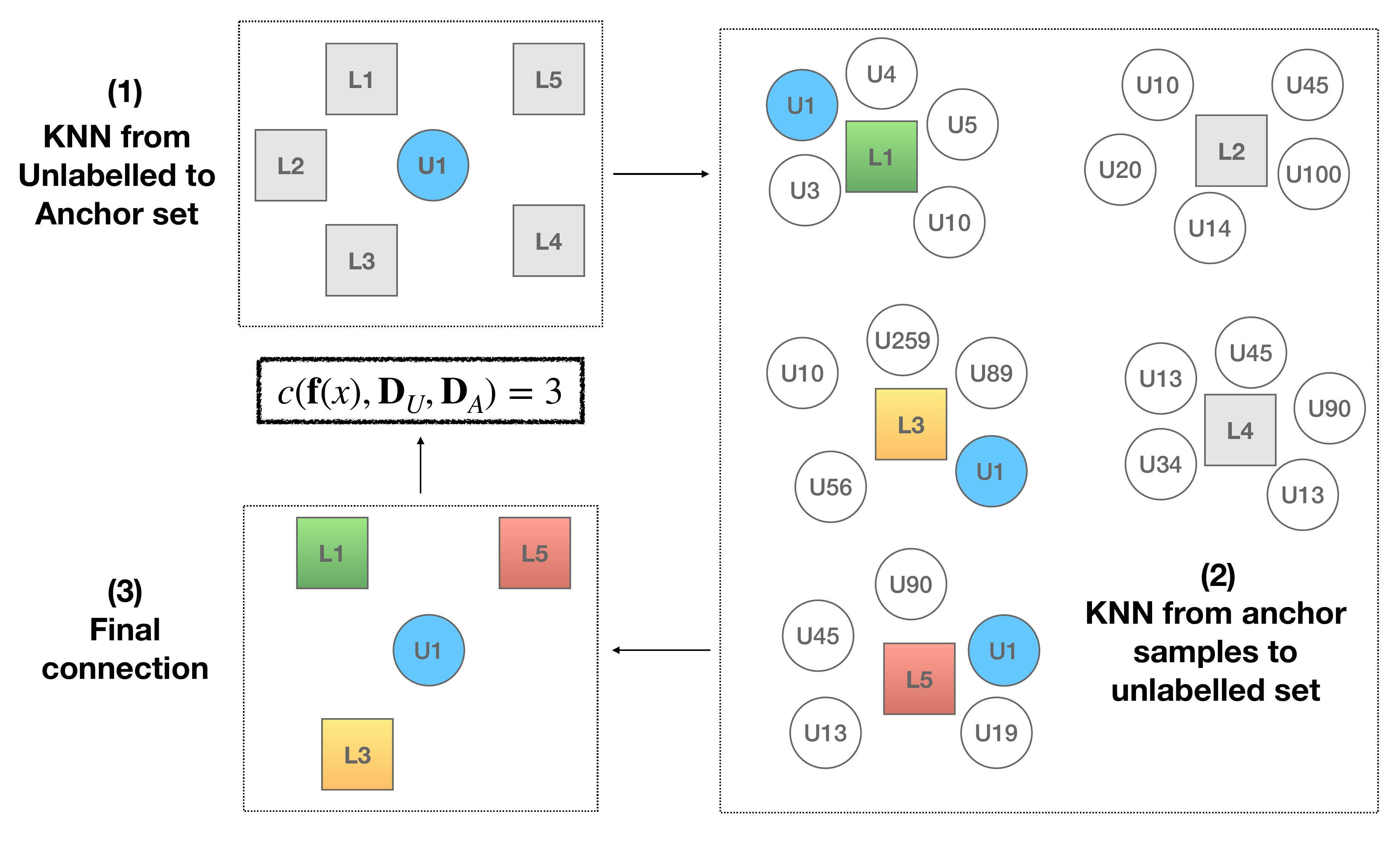}
    \caption{\textbf{ASP}: 1) find KNN samples from an informative unlabelled sample to the anchor set $\mathcal{D}_A$; 2)  find KNN samples from each anchor sample of (1) to the unlabelled set $\mathcal{D}_U$; and 3) calculate the number of surviving nearest neighbours. Samples with the smallest values of $c(.)$ are selected to be inserted into $\mathcal{D}_A$.
    }
    \label{fig:ASP}
\end{figure}
\label{sec:asp}

After estimating the pseudo label for informative unlabelled samples, we aim to update the anchor set with informative pseudo-labelled samples to maintain density score from~\eqref{eq:informativeness} accurate in later training stages. However, adding all pseudo-labelled samples will cause anchor set over-sized and increase hyper-parameter sensitivity. Thus, we propose the Anchor Set Purification (ASP) module to select the least connected pseudo-labelled samples to be inserted in the anchor set, as in (see Fig.~\ref{fig:ASP}):
\begin{equation}
    a(f_{\theta}(\mathbf{x}),\mathcal{D}_U,\mathcal{D}_A) =
\left\{\begin{array}{ll} 
1, & c(f_{\theta}(\mathbf{x}),\mathcal{D}_U,\mathcal{D}_A) \leq \alpha,\\
0, & \text{otherwise},
\end{array}
\right.	
\label{eq:anchor_set_insert}
\end{equation}
where the pseudo-labelled samples with $a(f_{\theta}(\mathbf{x}),\mathcal{D}_U,\mathcal{D}_A) = 1$ and  $\tilde{\mathbf{y}}=g(f_{\theta}(\mathbf{x}),\mathcal{D}_A)$ from~\eqref{eq:pseudo_label} are inserted into the anchor set. %
The information content $c(f_{\theta}(\mathbf{x}),\mathcal{D}_U,\mathcal{D}_A)$ of a pseudo-labelled sample $f_{\theta}(\mathbf{x})$ in~\eqref{eq:anchor_set_insert} is computed in three steps (see Fig.~\ref{fig:ASP}): 1) find the KNN samples
$\mathcal{N}(f_{\theta}(\mathbf{x}),\mathcal{D}_A)$ from $f_{\theta}(\mathbf{x})$ to the anchor set $\mathcal{D}_A$; 2) for each of the $K$ elements
$(\mathbf{x}_A,\mathbf{y}_A) \in \mathcal{N}(f_{\theta}(\mathbf{x}),\mathcal{D}_A)$, find the KNN set
$\mathcal{N}(f_{\theta}(\mathbf{x}_A),\mathcal{D}_U)$ from $f_{\theta}(\mathbf{x}_A)$ to the unlabelled set $\mathcal{D}_U$; and 3) $c(f_{\theta}(\mathbf{x}),\mathcal{D}_U,\mathcal{D}_A)$ is calculated to be the number of times that the pseudo-labelled sample $\mathbf{x}$ appears in the KNN sets $\mathcal{N}(f_{\theta}(\mathbf{x}_A),\mathcal{D}_U)$ for the $K$ elements of set $\mathcal{N}(f_{\theta}(\mathbf{x}),\mathcal{D}_A)$. The threshold $\alpha$ in~\eqref{eq:anchor_set_insert} is computed with $\alpha = \min_{\mathbf{x} \in \mathcal{D}_S}c(f_{\theta}(\mathbf{x}),\mathcal{D}_U,\mathcal{D}_A)$.

\section{Experiments}

\begin{table*}[t!]
\centering
\caption{Mean AUC testing set results over the 14 disease classes of Chest X-Ray14 for different labelled set training percentages. * indicates the methods that use DenseNet-169 as backbone architecture. \textbf{Bold} number means the best result per label percentage and \underline{underline} shows previous best results. %
}
\label{tab:main}
\scalebox{0.85}{
\begin{tabular}{c|l|c|c|c|c|c}
\toprule \hline 
Method Type & Label Percentage  & 2\%           & 5\%            & 10\%           & 15\%           & 20\%  \\ \hline

\multirow{3}{*}{Consistency based} & SRC-MT*   \cite{liu2020semi}         & 66.95         & 72.29          & 75.28          & 77.76          & 79.23 \\
                                  & NoTeacher \cite{unnikrishnan2020semi}  & 72.60 & 77.04 & 77.61 & N/A &79.49\\ 
                                 & S$^2$MTS$^2$ ~\cite{liu2021self} & \underline{74.69} & \underline{78.96} & \underline{79.90} & \underline{80.31}  &\underline{81.06}   \\ \hline
\multirow{2}{*}{Pseudo Label} & Graph XNet* \cite{aviles2019graphx}        & 53.00            & 58.00             & 63.00            & 68.00             & 78.00     \\
& UPS~\cite{rizve2020defense} &   65.51      & 73.18  &        76.84       &      78.90         &         79.92          \\              
& Ours     & \textbf{74.82} & \textbf{79.20} & \textbf{80.40} & \textbf{81.06} & \textbf{81.77}     \\ \hline  \bottomrule
\end{tabular}
}
\end{table*}

For the experiments below, we use the Chest X-Ray14~\cite{wang2017chestx} and ISIC2018~\cite{tschandl2018ham10000, codella2019skin} datasets. \\
\textbf{Chest X-Ray14} contains 112,120 CXR images from 30,805 different patients. There are 14 labels (each label is a disease) and No Finding class, where each patient can have multiple labels, forming a multi-label classification problem. To compare with previous papers~\cite{aviles2019graphx,liu2020semi}, we adopt the official train/test data split~\cite{wang2017chestx}. We report the classification result on the test set (26K samples) using area under the ROC curve (AUC), and the learning process uses training sets containing different proportions of the labelled data in  $\{2\%,5\%,10\%,15\%,20\%\}$. \\
\textbf{ISIC2018} is a skin lesion dataset that contains 10,015 images with seven  labels. Each image is associated with one of the labels, forming a multi-class classification problem. 
We follow the train/test split from~\cite{liu2020semi} for fair comparison, where the training set contains 20\% of labelled samples and 80\% of unlabelled samples. 
We report the AUC, Sensitivity, and F1 score results.

\begin{table*}[t!]
\centering
\caption{Class-level AUC testing set results comparison between our approach and other semi-supervised SOTA approaches trained with \textbf{20\%} of labelled data on Chest Xray-14. * denotes the models use DenseNet-169 as backbone. \textbf{Bold} number means the best result per class and \underline{underlined} shows second best results.}
\label{tab:semi-results}
\scalebox{0.85}{
\begin{tabular}{@{}c|c|c|c|c|c|c|c@{}}
\toprule \hline 
Method Type & Supervised & \multicolumn{3}{c|}{Consistency based} &  \multicolumn{3}{c|}{Pseudo-labelling} \\ \hline \hline 
Method          & Densenet-121                           & MT ~\cite{tarvainen2017mean} *  & SRC-MT ~\cite{liu2020semi} *  & S$^2$MTS$^2$ ~\cite{liu2021self} & GraphXNet ~\cite{aviles2019graphx} & UPS~\cite{rizve2020defense} & Ours  \\ \hline  \hline 
Atelectasis     &     75.75                              & 75.12            &     75.38                 & \underline{78.57}           &       71.89 & 77.09 &  \textbf{79.53}      \\
Cardiomegaly    &     80.71                               & 87.37            &     87.7                & \underline{88.08}            &       87.99   & 85.73 &  \textbf{89.03}   \\
Effusion        &     79.87                              & 80.81             &       81.58              & \underline{82.87}           &         79.2   &  81.35&  \textbf{83.56} \\
Infiltration    &       69.16                           & 70.67            &       70.4               & 70.68            &          \textbf{72.05}  & 70.82 &  \underline{71.40} \\
Mass            &      78.40                             & 77.72             &       78.03               & \textbf{82.57}           &      80.9      &  81.82&  \underline{82.49} \\                                                                                 
Nodule          &     74.49                                & 73.27            &      73.64              & \underline{76.60}           &       71.13   & 76.34 &\textbf{77.73}  \\
Pneumonia       &     69.55                                  & 69.17            &     69.27               & 72.25           &      \textbf{76.64}     & 70.96 &   \underline{73.86}  \\
Pneumothorax    &       84.70                            & 85.63             &       86.12               &  \underline{86.55}           &        83.7   & 85.86 & \textbf{86.95}  \\
Consolidation   &      71.85                             & 72.51            &       73.11              & \underline{75.47}          &         73.36    &  74.35& \textbf{75.50}  \\
Edema           &      81.61                              & 82.72             &      82.94               &\underline{84.83}           &       80.2   & 83.56 &\textbf{84.95}  \\
Emphysema       &    89.75                               & 88.16            &      88.98              & \underline{91.88}            &         84.07  & 91.00 &\textbf{93.36} \\
Fibrosis        &      79.30                              & 78.24           &     79.22              & \underline{ 81.73 }           &        80.34 & 80.87 &\textbf{81.86}  \\
Pleural Thicken &       73.46                             & 74.43            &     75.63               & \underline{  76.86 }           &       75.7  & 75.55 &\textbf{77.60} \\
Hernia          &      86.05                             & \textbf{87.74 }   &       \underline{87.27}           & 85.98           &        87.22   & 85.62&85.89  \\ \hline 
Mean            &     78.19                              & 78.83            &      79.23              & \underline{81.06}           &        78.00  & 79.92 & \textbf{81.77}    \\ \hline \bottomrule
\end{tabular}
}%
\end{table*}

\subsection{Implementation Details}

For both datasets, we use DenseNet-121~\cite{huang2017densely} as our backbone model. 
For Chest X-Ray14, 
the dataset pre-processing consists of resizing the images to 512 $\times$ 512 for faster processing.
For the optimisation, we use Adam optimizer~\cite{kingma2014adam}, batch size 16 and learning rate 0.05. 
During training, we use data augmentation based on random crop and resize, and random horizontal flip.
We first train 20 epochs on the initial labelled subset to warm-up the model for feature extraction. 
Then we train for 50 epochs, where in every 10 epochs we update the anchor set with ASP from Sec.~\ref{sec:asp}.
For the KNN classifier in~\eqref{eq:information_estimator}, we set K to be 200 for 2\% and 5\% (of labelled data) and 50 for remaining label proportions. 
These values are set based on validation results, but our approach is robust to a large range K values -- we show an ablation study that compares the performance of our method for different values of K.
For ISIC2018, we resize the image to 224 $\times$ 224 for fair comparison with baselines.
For the optimisation, we use Adam optimizer~\cite{kingma2014adam}, batch size 32 and learning rate 0.001. 
During training, data augmentation is also based on random crop and resize, and random horizontal flip.
We warm-up the model for 40 epochs and then we train for 100 epochs, where in every 20 epochs, we update the anchor set with ASP. For the KNN classifier, K is set to 100 based on validation set.
The code is written in Pytorch~\cite{NEURIPS2019_9015} and we use two RTX 2080ti Gpus for all experiments. KNN computation takes 5 sec for Chest X-ray14 unlabelled samples with Faiss~\cite{JDH17} library for faster processing.
We follow~\cite{liu2020semi, liu2021self,tarvainen2017mean} to maintain an exponential moving average (EMA) version of the trained model, which is only used for evaluation not for training.

\subsection{Thorax Disease Classification Result}
\label{sec:chestxray14_results}

For the results on Chest X-Ray14 in Table~\ref{tab:main}, 
our method, 
NoTeacher~\cite{unnikrishnan2020semi}, UPS~\cite{rizve2020defense}, and S$^2$MTS$^2$~\cite{liu2021self} use the DenseNet-121 backbone, 
while SRC-MT~\cite{liu2020semi} 
and GraphXNet~\cite{aviles2019graphx} use DenseNet-169~\cite{huang2017densely}.
SRC-MT~\cite{liu2020semi} is a consistency-based SSL; NoTeacher~\cite{unnikrishnan2020semi} extends MT by replacing the EMA process with two networks combined with a probabilistic graph model; %
S$^2$MTS$^2$~\cite{liu2021self} combines self-supervised pre-training with MT fine-tuning; and
GraphXNet~\cite{aviles2019graphx} constructs a graph from dataset samples and assigns pseudo labels to unlabelled samples through label propagation; and 
UPS~\cite{rizve2020defense} applies probability  and uncertainty thresholds to enable the pseudo labelling of unlabelled samples.
All methods use the official test set~\cite{wang2017chestx}.
Our approach achieves the SOTA results for all percentages of training labels. 
Compared to the pseudo-labelling approaches UPS and GraphXNet, our approach outperforms them by a margin between 3\% to 20\%. Compared to the consistency-based approaches SRC-MT and NoTeacher, our method consistently achieves 2\% improvement for all cases, even though we use a backbone architecture of lower capacity (i.e., DenseNet-121 instead of DenseNet-169). 
Compared with the previous SOTA, our method outperforms S$^2$MTS$^2$~\cite{liu2021self} by 1\% AUC in all cases, which is remarkable because our method is initialised with an ImageNet pre-trained model instead of an expensive self-supervised pre-training approach.

The class-level performances using 20\% of the labelled data of SSL methods are shown in Table~\ref{tab:semi-results}, which demonstrates that our method achieves the best result in 10 out of the 14 classes.
Our method surpasses the previous pseudo-labelling method GraphXNet by 3.7\% and threshold based pseudo-labelling method~\cite{rizve2020defense} by 1.8\%. Our method also outperforms consistency-based methods MT~\cite{tarvainen2017mean} and SRC-MT~\cite{liu2020semi} by more than 2\%. For method S$^2$MTS$^2$~\cite{liu2021self} with self-supervised learning, our method can outperform it using an ImageNet pre-trained model, alleviating the need of a computationally expensive self-supervised pre-training.

\subsection{Skin Lesion Classification Result}
\label{sec:isic_results}

\begin{table}[t!]
      \centering
      \caption{AUC, Sensitivity and F1 testing results on  ISIC2018, where 20\% of the training set is labelled. \textbf{Bold} shows the best result per measure, and \underline{underline} shows second best results. %
      }
      \scalebox{0.9}{
      \begin{tabular}{c|c|c|c}
      \toprule Method     & AUC   & Sensitivity & F1    \\ \hline
                Supervised & 90.15 & 65.50       & 52.03 \\ \hline
                SS-DCGAN ~\cite{diaz2019retinal}                         & 91.28 & 67.72 & 54.10 \\ \hline
                TCSE~\cite{li2018semi}                             & 92.24 & 68.17 & 58.44 \\ 
                TE       ~\cite{laine2016temporal}                        & 92.70 & 69.81 & 59.33\\ 
                MT ~\cite{tarvainen2017mean}        & 92.96 & 69.75       & 59.10 \\ 
                SRC-MT ~\cite{liu2020semi}     & \underline{93.58} & \underline{71.47}       & \underline{60.68} \\ \hline
                Self-training ~\cite{bai2017semi} & 90.58 & 67.63 & 54.51 \\    
                Ours       & \textbf{94.36} & \textbf{72.14}       & \textbf{62.23} \\ \hline \bottomrule
      \end{tabular}}
      \label{tab:isic}
  \end{table}
 
 We show the results on ISIC2018 in Table~\ref{tab:isic}, where competing methods are based on self-training~\cite{bai2017semi}, generative adversarial network (GAN) to augment the labelled set~\cite{diaz2019retinal}, temporal ensembling~\cite{laine2016temporal},  MT~\cite{tarvainen2017mean} and its extension~\cite{liu2020semi}, and a  DenseNet-121~\cite{huang2017densely} baseline trained with 20\% of the training set. 
 Compared with consistency-based  approaches~\cite{tarvainen2017mean,liu2020semi,li2018semi}, 
 our method improves between 0.7\% and 3\% in AUC and around 1\% in F1 score. 
 Our method also outperforms  previous self-training approach~\cite{bai2017semi} by a large margin in all measures.

\begin{figure}
    \centering
    \begin{subfigure}[t]{0.49\linewidth}
        \centering
        \includegraphics[width=\linewidth,page=12]{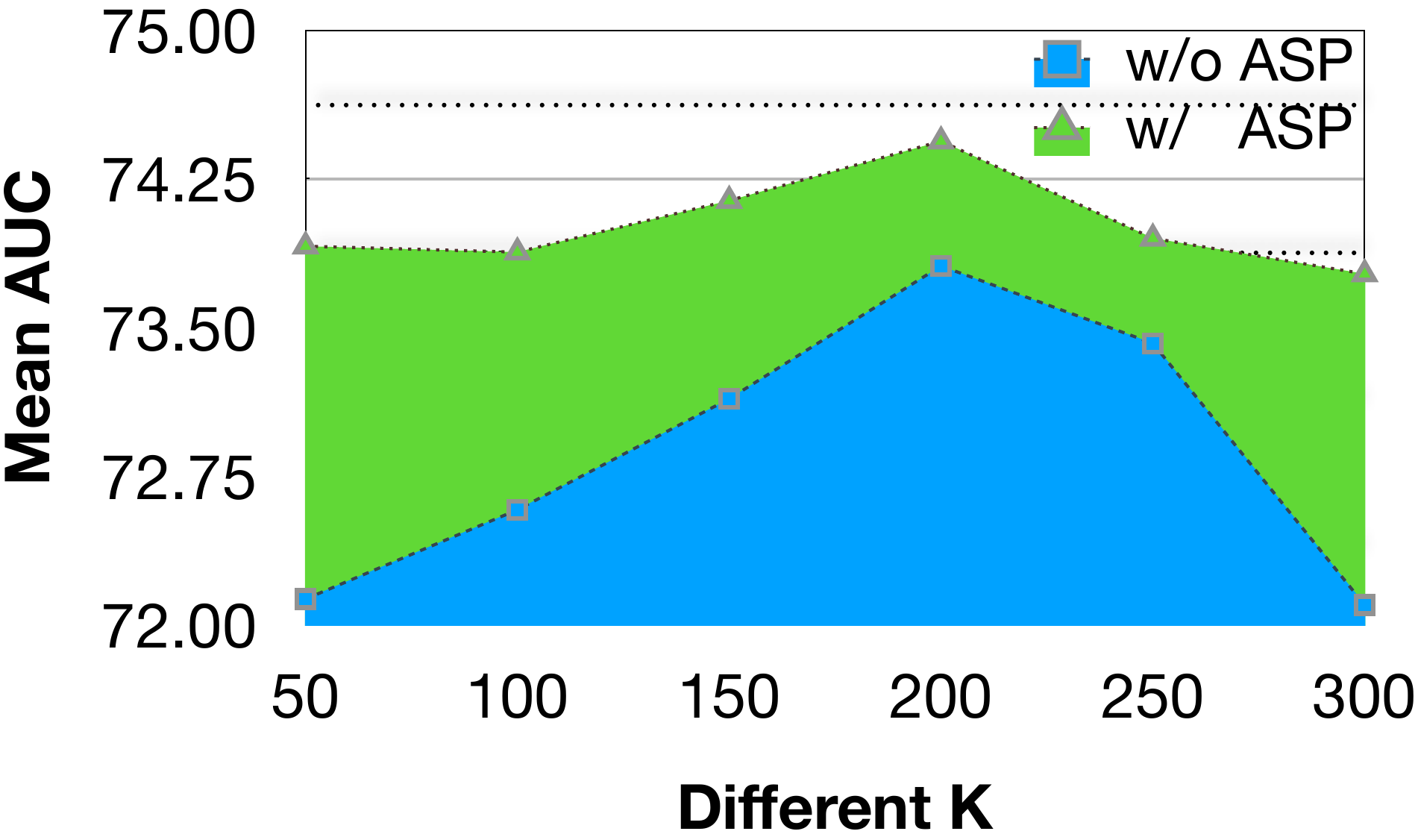}
    \end{subfigure}%
    ~
    \begin{subfigure}[t]{0.49\linewidth}
        \centering
        \includegraphics[width=\linewidth,page=13]{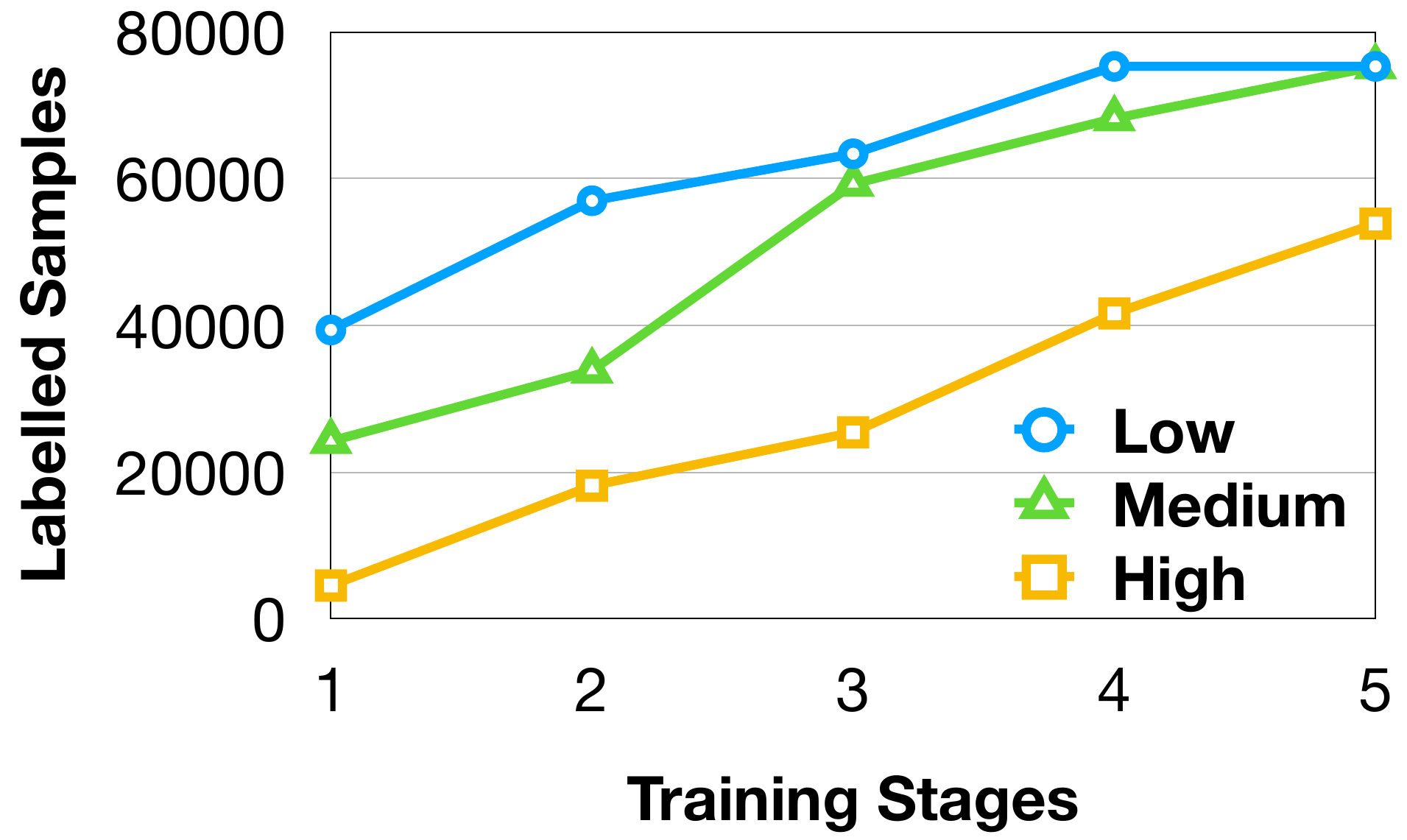}
    \end{subfigure}

    \caption{(Left) Mean AUC testing results for different values for K in the KNN (for CDSI in~\eqref{eq:informativeness} and pseudo-labelling in~\eqref{eq:pseudo_label_types}), where the green region uses ASP and blue region does not use ASP.
    (Right) Mean size of $\mathcal{D}_L$ at every training stage when adding unlabelled samples of high, medium and low information content according to~\eqref{eq:information_estimator}. 
        Model is trained on Chest X-Ray14, where 2\% of the training is labelled.
        } %
    \label{fig:KASP_label_size}
\end{figure}

\subsection{Ablation Study}

\begin{table}[t!]
\centering
\caption{Ablation study on Chest X-ray14 (2\% labelled). 
    Starting with a baseline classifier (DenseNet-121), we test the selection of unlabelled samples (to be provided with a pseudo-label) with different information content, according to~\eqref{eq:information_estimator} (i.e., low, medium, high), and the use of the anchor set purification (ASP) module.
    }
\label{tab:AC_ASP}
\scalebox{0.8}{
\begin{tabular}{cc|c}
\toprule \hline
\multicolumn{1}{c|}{Information Content}     & ASP & AUC $\pm$ std    \\ \hline
\multicolumn{2}{c|}{Baseline}                                &   65.84 $\pm$ 0.14 \\ \hline
\multicolumn{1}{c|}{\multirow{2}{*}{Low}}    & \textcolor{red}{\XSolidBrush}     &   67.18 $\pm$ 2.40               \\ \cline{2-3} 
\multicolumn{1}{c|}{}                        & \textcolor{green}{\Checkmark}   &   67.76 $\pm$ 1.05               \\ \hline
\multicolumn{1}{c|}{\multirow{2}{*}{Medium}} &  \textcolor{red}{\XSolidBrush}     &   70.83 $\pm$ 1.49               \\ \cline{2-3} 
\multicolumn{1}{c|}{}                        & \textcolor{green}{\Checkmark}    &    71.16 $\pm$ 0.51              \\ \hline
\multicolumn{1}{c|}{\multirow{2}{*}{High}}   & \textcolor{red}{\XSolidBrush}     &     73.81 $\pm$ 0.75             \\ \cline{2-3} 
\multicolumn{1}{c|}{}                        & \textcolor{green}{\Checkmark}    &    \textbf{74.44 $\pm$ 0.38}              \\  \hline \bottomrule
\end{tabular}
}
\end{table}

For the ablation study, we test each of our three contributions and visualize the data distribution of selected subset with high and low informative samples on the Chest X-Ray14~\cite{wang2017chestx} with 2\% labelled training set, where for CDSI and ASP, we run each experiment three times and show the mean and standard deviation of the AUC results.\\
\textbf{Cross-distribution Sample Informativeness (CDSI).} We first study in Table~\ref{tab:AC_ASP} how performance is affected by pseudo-labelling unlabelled samples with different degrees of informativeness (low, medium and high) using our CDSI.
Starting from the baseline classifier DenseNet-121 that reaches an AUC of 65\%, we observe that pseudo-labelling low-information content unlabelled samples yields the worst result (around 67\% AUC) and selecting high-information content unlabelled samples produces the best result (around 73\% AUC). 
Figure~\ref{fig:KASP_label_size} (right) plots how the size of the labelled set $\mathcal{D}_L$ during training depends on the degree of informativeness of the unlabelled samples to be pseudo-labelled.
These results show that: 1) unlabelled samples of high-information content enables the construction of a smaller labelled set (compared with unlabelled samples of low- or medium-information content), allowing a more efficient training process that produces a more accurate KNN classifier; and 2) the standard deviation of the results in Table~\ref{tab:AC_ASP} are smaller 
when selecting the unlabelled samples of high-information content, compared with the low- or medium-information content. This second point can be explained by the class imbalance issue in Chest X-Ray14, where the selection of low-information content samples will enable the training of majority classes, possibly producing an ineffective training for the minority classes that can increase the variance in the results. \\
\noindent\textbf{Anchor Set Purification (ASP)}. Also in Table~\ref{tab:AC_ASP}, we compare ASP with an alternative method that selects all pseudo-labelled samples to be included into the anchor set for the low-, medium- and high-information content unlabelled samples.
Results show that the ASP module improves AUC between 0.3\% and 1.0\% and reduces standard deviation between 0.4\% and 1.4\%. 
This demonstrates empirically that the ASP module enables the formation of a more informative anchor set that improves the pseudo-labelling accuracy, and consequently the final AUC results.
Furthermore, in Figure~\ref{fig:KASP_label_size} (left), ASP is shown to stabilise the performance of the method with respect to $K \in \{50,100,150,200,250,300\}$ for the KNN classifier of~\eqref{eq:informativeness}.
In particular, with ASP, the difference between the best and worst AUC results is around 1\%, while without ASP, the difference grows to 2\%.
This can be explained by the fact that without ASP, the anchor set grows quickly with relatively less informative pseudo-labelled samples, which reduces the stability of the method. 

\begin{table}[t!]
\centering
\caption{AUC testing set results on Chest X-ray14 (2\% labelled) for different pseudo labelling strategies  ($\alpha$ denotes the linear coefficient combining the model and KNN predictions).}
\scalebox{0.8}{
\begin{tabular}{lc|c}
\toprule \hline
\multicolumn{1}{c|}{Pseudo-label Strategies}                       & Methods                           & AUC                               \\ \hline
\multicolumn{1}{c|}{Baseline}                           & -                                                 & 65.84                             \\ \hline
\multicolumn{1}{c|}{\multirow{2}{*}{Single Prediction}} & Model prediction                                  & 72.63                             \\
\multicolumn{1}{l|}{}                                   & KNN prediction                                    & 72.45                             \\ \hline
\multicolumn{1}{c|}{\multirow{2}{*}{Mixup}}             & random sampled $\alpha$                           & 73.23                             \\
\multicolumn{1}{l|}{}                                   & MixUp~\cite{zhang2017MixUp}                      & 69.28                             \\ \hline
\multicolumn{1}{c|}{Ours}                               & Informative Mixup                                 & \textbf{74.44}        \\ \hline \bottomrule
\end{tabular}}
\label{tab:density_MixUp}
\end{table}

\vspace{-0.07cm}
\noindent \textbf{Informative Mixup (IM)} In Table~\ref{tab:density_MixUp}, we show that our proposed IM in~\eqref{eq:pseudo_label} produces a more accurate pseudo-label, where we compare it with alternative pseudo-label methods, such as with only the model prediction, only the KNN prediction, random sample $\alpha$ from beta distribution to replace the density score in~\eqref{eq:pseudo_label}, and regular MixUp~\cite{zhang2017MixUp}. 
It is clear that the use of model or KNN predictions as pseudo labels does not work well most likely because of confirmation bias (former case) or the inaccuracy of the KNN classifier (latter case).
MixUp~\cite{zhang2017MixUp} does not show good accuracy either, as also observed in~\cite{wang2019baseline} and~\cite{kim2021co}, when MixUp is performed in multi-label images or multiple single-object images. The random sampling of $\alpha$ for replacing density score shows a better result than MixUp, but the lack of an image-based weight to balance the two predictions, like in~\eqref{eq:pseudo_label}, damages performance.
Our proposed IM shows a result that is at least 1.5\% better than any of the other pseudo-labelling approaches, showing the importance of using the density of the unlabelled sample in the anchor set to weight the contribution of the model and KNN classifiers.

\begin{figure}
    \centering
    \includegraphics[clip, trim=7.0cm 14cm 5.0cm 0.0cm,width=0.9\linewidth,page=10]{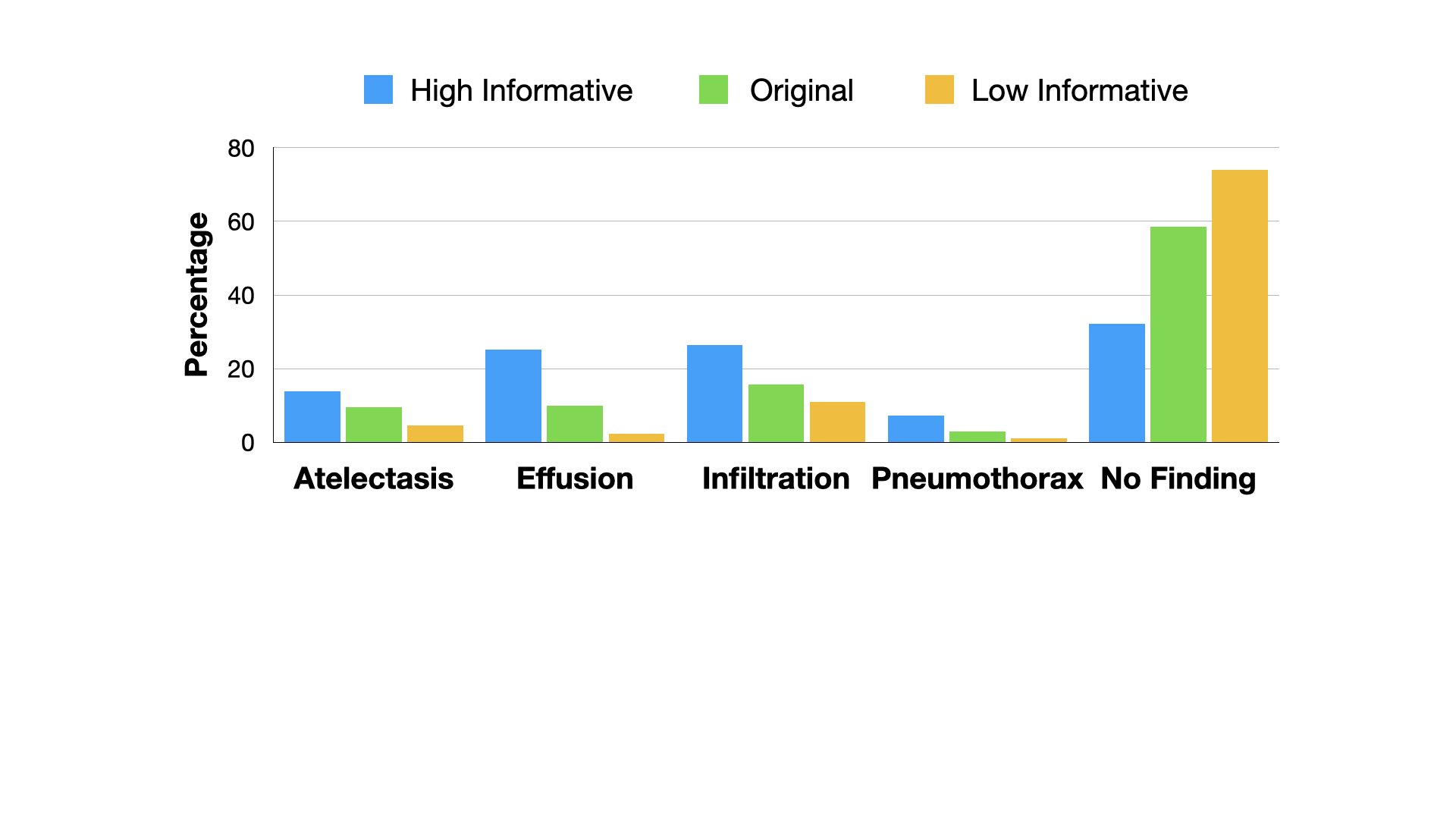}
    \caption{The selection of highly informative unlabelled samples (blue) promote a more balanced 
    learning process, where the difference in the number of samples belonging to the minority or majority classes is smaller
    than if we selected unlabelled samples with low informativeness (yellow). Green shows the original data distribution]. Full 14-class distributions are shown in the supplementary material. 
    }
    \label{fig:data_distribution}
\end{figure}
 
 \noindent The \textbf{imbalanced learning mitigation} is studied in Figure~\ref{fig:data_distribution}, which shows the histogram of label distribution in percentage (for a subset of four disease minority classes and the No Finding majority class) by selecting unlabelled samples of high (blue) and low (yellow) information content. We also show the original label distribution in green for reference.  
 
 Notice that the selection of highly informative samples significantly increases the percentage of disease minority classes (from between 5\% and 10\% to almost 30\%) and decreases the percentage of the No Finding majority class (from 60\% to 30\%), creating a more balanced distribution of these five classes. 
 This indicates that our informative sample selection can help to mitigate the issue of imbalanced learning.
We include the full 14-classes histograms in the supplementary material.

\section{Discussion and Conclusion}
In this work, we introduced the anti-curriculum pseudo-labelling (ACPL) SSL method. Unlike traditional pseudo-labelling methods that use a threshold to select confidently classified samples, ACPL uses a new mechanism to select highly informative unlabelled samples for pseudo-labelling and an ensemble of classifiers to produce accurate pseudo-labels. This enables ACPL to address MIA multi-class and multi-label imbalanced classification problems. 
We show in the experiments that ACPL outperforms previous consistency-based, pseudo-label based and self-supervised SSL methods in multi-label Chest X-ray14 and multi-class ISIC2018 benchmarks. We demonstrate in the ablation study the influence of each of our contributions and we also show how our new selection of informative samples addresses MIA imbalanced classification problems.
For future work, it is conceivable that ACPL can be applied  can be applied to more general computer vision problems, so we plan to test ACPL in traditional computer vision benchmarks. We would also explore semi-supervised classification with out-of-distribution (OOD) data in the initial labelled and unlabelled sets as our method currently assume all samples are in-distribution.

{\small
\bibliographystyle{ieee_fullname}
\bibliography{egbib}
}

\end{document}


\title{Supplementary Material for \\ACPL: Anti-curriculum Pseudo-labelling for Semi-supervised Medical Image Classification}

\author{
\parbox{0.7\linewidth}{\centering $\quad$ Fengbei Liu\textsuperscript{\rm 1}\thanks{First two authors contributed equally to this work.}  $\quad$ Yu Tian\textsuperscript{\rm 1}\footnotemark[1] $\quad$    Yuanhong Chen\textsuperscript{\rm 1} $\quad$ Yuyuan Liu\textsuperscript{\rm 1}$\newline$  $\quad$  Vasileios Belagiannis\textsuperscript{\rm 2} $\quad$ $\quad$ $\quad$ Gustavo Carneiro\textsuperscript{\rm 1} $\newline$   
\textsuperscript{\rm 1} Australian Institute for Machine Learning, University of Adelaide \\
\textsuperscript{\rm 2} Universit\"at Ulm, Germany} 
}
\maketitle

\section{Additional Ablation study}
\begin{table}[t!]
\centering
\caption{Ablation study of the number of information content sets in Eq.2 (2, 3, 4 sets) with model training performance (in terms of mean AUC testing set results) and number of training stages with 2\% and 20\% labelled set on Chest X-ray14~\cite{wang2017chestx}.}
\scalebox{0.85}{
\begin{tabular}{c|c|c|c}
\toprule \hline
Number of Inform. Cont. Sets in Eq.2       & 2 & 3 & 4                        \\ \hline
Number of Training Stages & 5 & 5 & 9   \\ \hline
2\%           & 71.28 & 74.44 & 74.37                         \\ \hline
20\%            &  79.56 & 81.51 & 81.60                       \\ \hline \bottomrule 
\end{tabular}
}
\label{tab:eq2sets}
\end{table}

 \textbf{The Number of Information Content Sets in Eq.2} is studied in Table~\ref{tab:eq2sets}, which shows the model training performance (in terms of mean AUC testing set results) and number of training stages using 2\% and 20\% labelled set on Chest X-ray14~\cite{wang2017chestx}. The default setting used in the paper is to have three information content sets, namely {low, medium, high}. As shown in Table~\ref{tab:eq2sets}, the selection of only two sets produces the worst results because the pseudo-labelled set becomes less informative and imbalanced. The selection of four sets produces similar results as with three sets. However, with this additional set, the number of new pseudo labelled samples are greatly reduced for every training stage, forcing the number of training stages to grow.  Hence, by selecting three sets we reach a good balance between training time and accuracy.
 
\section{Data Distribution}

\begin{figure*}
    \centering
    \includegraphics[page=6,width=\linewidth]{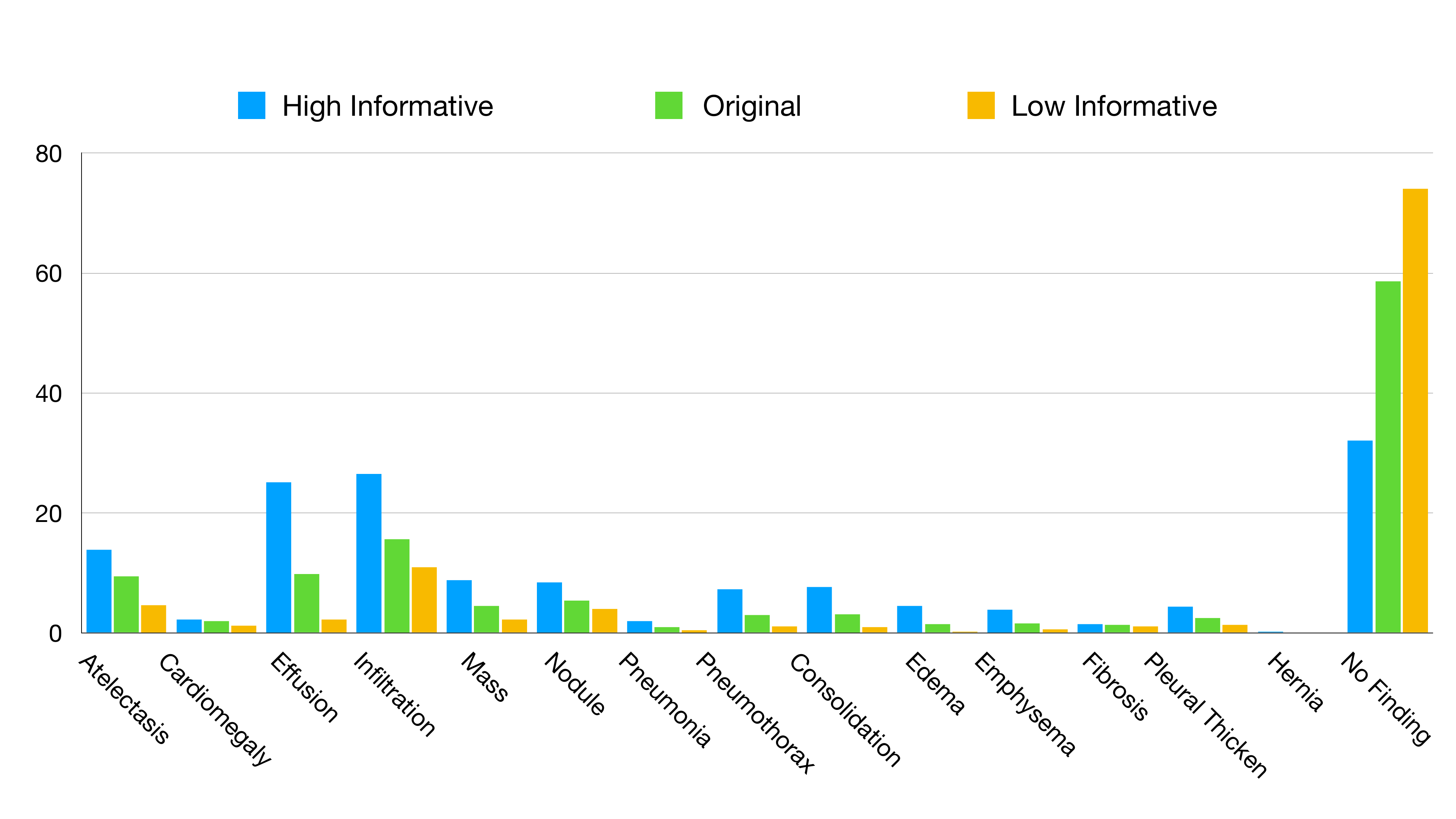}
    \caption{Histogram of label distribution in percentage of all 14 classes from Chest X-ray14 plus the class 'No Finding'. 
    Blue for high information content subset and yellow for low information content subset. Green is the original data distribution. }
    
    \label{fig:full_data_distribution}
\end{figure*}

In Figure~\ref{fig:full_data_distribution}, we show the data distribution of all classes of Chest X-ray14 (plus the class 'No Findings')~\cite{wang2017chestx}. Notice that the selection of high information content samples (blue) creates a more balanced distribution compared with the selection of low information content (yellow) or the original data distribution (green).

\section{Visualization of Classification Results}

\begin{figure*}
    \centering
    \includegraphics[clip, trim=0.0cm 12cm 1.0cm 0.0cm,width=\linewidth,page=6]{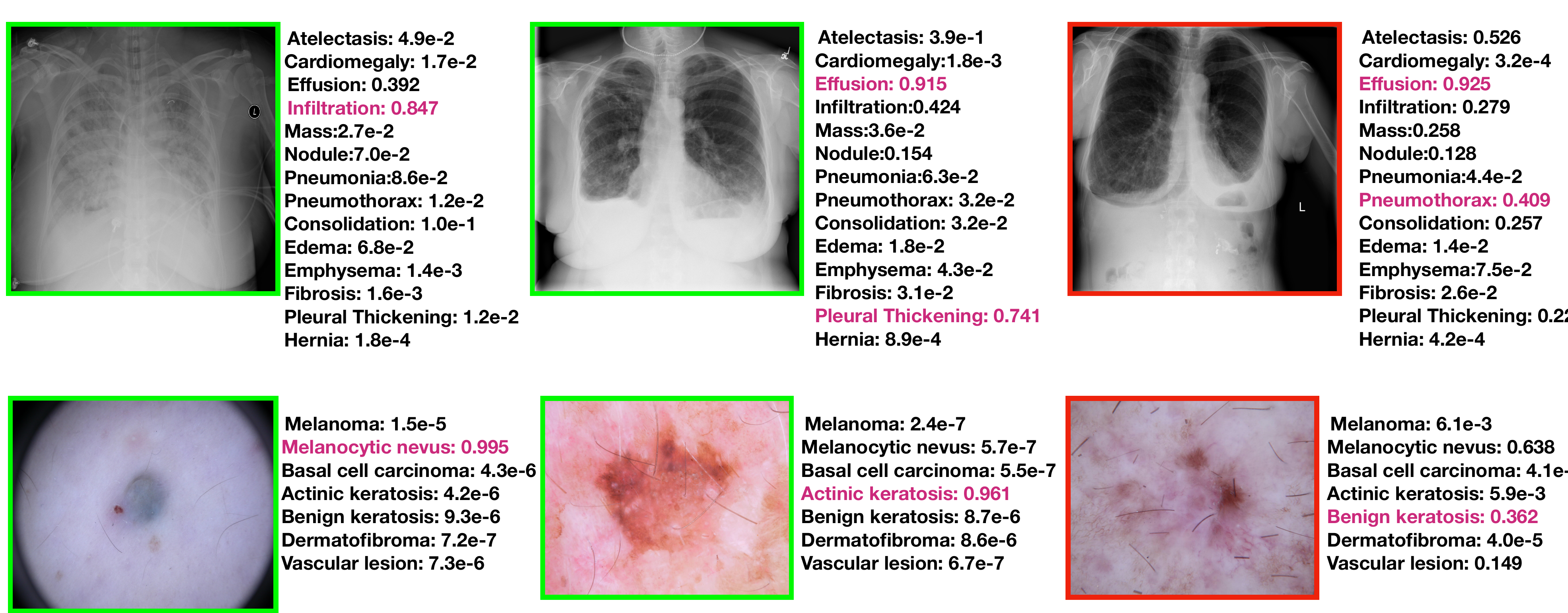}
    \caption{Pseudo-labelling of high-information content unlabelled samples estimated with the \textbf{density mixup} prediction for Chest Xray-14~\cite{wang2017chestx} (top) and ISIC2018~\cite{tschandl2018ham10000} (bottom) datasets. Green border denotes accurate prediction and red border represents inaccurate prediction. Classes with red color represent the ground truth. %
    }
    \label{fig:visual_results}
\end{figure*}

Figure~\ref{fig:visual_results} shows examples of pseudo-labels produced by our density mixup for both Chest Xray-14~\cite{wang2017chestx} (top) and ISIC2018~\cite{tschandl2018ham10000} (bottom) datasets.
 
{\small
\bibliographystyle{ieee_fullname}
\bibliography{egbib}
}